\pdfoutput=1

\documentclass[11pt]{article}

\usepackage{acl}
\usepackage{times}
\usepackage{latexsym}
\usepackage{amsmath}
\usepackage{subcaption}

\usepackage[T1]{fontenc}

\usepackage[utf8]{inputenc}
\usepackage{microtype}

\usepackage{inconsolata}

\usepackage{graphicx}

\usepackage[ruled,vlined]{algorithm2e}

\usepackage{amsmath}
\usepackage{amssymb}

\usepackage{natbib}
\usepackage{amsthm}

\title{HoPE: Hyperbolic Rotary Positional Encoding for Stable Long-Range Dependency Modeling in Large Language Models}

\author{
  Chang Dai\textsuperscript{1}  \quad
  Hongyu Shan\textsuperscript{2}  \quad
  Mingyang Song\textsuperscript{3} \quad
  Di liang\textsuperscript{4} \quad \\
  \textsuperscript{1}Peking University 
  \textsuperscript{2}Tianjin University \\
  \textsuperscript{3}Tencent 
  \textsuperscript{4}Fudan University \\
  \texttt{daichang@pku.edu.cn}, 
  \texttt{shhy@tju.edu.cn}, \\
  \texttt{nickmysong@tencent.com}, 
  \texttt{dliang@fudan.edu.cn}
}

\begin{document}

\maketitle
\begin{abstract}
Positional encoding mechanisms enable Transformers to model sequential structure and long-range dependencies in text. While absolute positional encodings struggle with extrapolation to longer sequences due to fixed positional representations, and relative approaches like Alibi exhibit performance degradation on extremely long contexts, the widely-used Rotary Positional Encoding (RoPE) introduces oscillatory attention patterns that hinder stable long-distance dependency modelling. We address these limitations through a geometric reformulation of positional encoding. Drawing inspiration from Lorentz transformations in hyperbolic geometry, we propose Hyperbolic Rotary Positional Encoding (\textbf{HoPE}), which leverages hyperbolic functions to implement Lorentz rotations on token representations. Theoretical analysis demonstrates that RoPE is a special case of our generalized formulation. HoPE fundamentally resolves RoPE's slation issues by enforcing monotonic decay of attention weights with increasing token distances. Extensive experimental results, including perplexity evaluations under several extended sequence benchmarks, show that HoPE consistently exceeds existing positional encoding methods. These findings underscore HoPE's enhanced capacity for representing and generalizing long-range dependencies. Data and code will be available.
\end{abstract}

\section{Introduction}
Positional encoding mechanisms strive to provide Transformers ~\cite{vaswani2023attentionneed} with stable and expressive representations of the sequential structure, thereby addressing the order-agnostic nature of the multi-head attention module~\cite{raffel2023exploringlimitstransferlearning,anil2022exploringlengthgeneralizationlarge}. By encoding information about the relative\cite{shaw2018selfattentionrelativepositionrepresentations} or absolute positions of tokens, positional encodings enable models to capture the intricacies of syntactic and semantic dependencies across different spans of text\cite{wang2020positionembeddingslearnempirical}. Without such positional signals, Transformers can struggle to fully delineate word-order information and effectively leverage long-range context\cite{haviv2022transformerlanguagemodelspositional}. A variety of strategies~\cite{ruoss2023randomizedpositionalencodingsboost,kazemnejad2023impactpositionalencodinglength,li2024functionalinterpolationrelativepositions,chen2023extendingcontextwindowlarge,xiong2023effectivelongcontextscalingfoundation} have been proposed to incorporate positional knowledge in large language models (LLMs), aiming to ensure reliable and generalizable sequence representations\cite{chowdhury2023monotoniclocationattentionlength,sun2022lengthextrapolatabletransformer}.

\begin{figure}[t]
    \centering
    \includegraphics[width=0.48\textwidth]{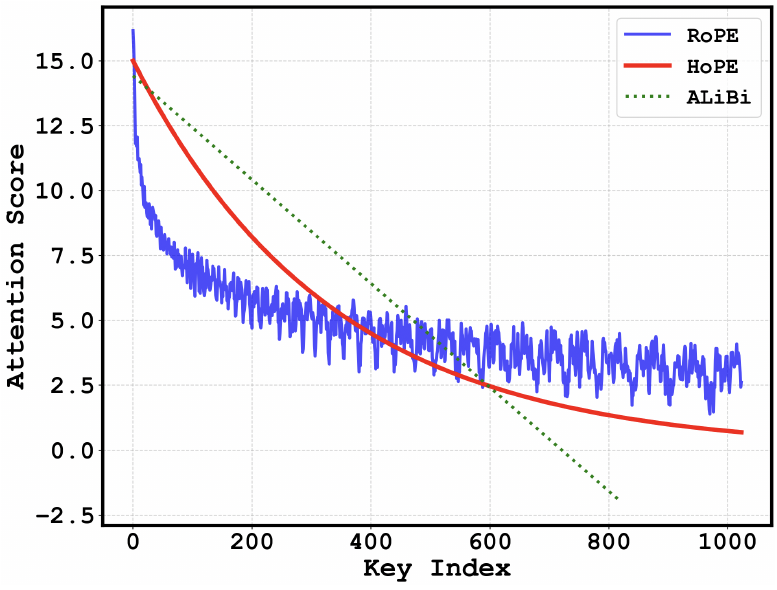} 
    \caption{Illustration of attention scores. For the same embedding, when using the RoPE, alibi, and HoPE methods to represent positional information, this shows the corresponding changing trends of the Attention scores as the distance varies.}
    \label{fig:example}
\end{figure}

 Absolute positional encodings \cite{devlin2019bertpretrainingdeepbidirectional}, which typically employ sinusoidal signals or learnable embeddings indexed by token positions, are straightforward to implement but often struggle with length extrapolation, as their fixed position representations do not naturally extend to unseen sequence lengths. In contrast, some approaches ~\cite{press2022trainshorttestlong,chi2022kerple,chi2023dissecting}, which introduce a distance-based attention bias, demonstrate improved performance over absolute encodings but can still degrade when sequences become very long, revealing limitations in capturing stable correlations at distant positions. More recently, Rotary Positional Encoding (RoPE)~\cite{su2023roformerenhancedtransformerrotary} has gained substantial traction by rotating query and key vectors at various frequencies as a function of token positions\cite{barbero2024roundroundgomakes}. However, RoPE exhibits an oscillatory attention pattern, in which attention weights fluctuate rather than decrease smoothly as the token distance increases, making it difficult to reliably represent long-distance dependencies, as shown in figure \ref{fig:example}.


Recent applications demonstrate that RoPE is widely adopted in many state‐of‐the‐art large-scale models, such as Llama, Gemini, and DeepSeek (\cite{}). Moreover, numerous efforts have been made to address the inherent limitations of RoPE (\cite{}). However, these approaches often rely on augmenting the original RoPE formulation with additional components or interpolation strategies to mitigate issues related to length extrapolation and representation learning, rather than revisiting and revising the core kernel design of RoPE itself. To directly tackle the distance noise problem illustrated in figure \ref{fig:example}, we draw inspiration from Lorentz transformations \cite{hall2000elementaryintroductiongroupsrepresentations}. Building on the observation that RoPE can be viewed as a specific case within the broader family of Lorentz transformations, we propose a novel positional encoding scheme. Specifically, we introduce Hyperbolic Rotary Positional Encoding (HoPE), which leverages hyperbolic sine and cosine functions to rotate query and key vectors. The resulting formulation ensures that attention weights decay monotonically with increasing token distance, thereby providing a more stable and robust representation for long-range dependencies.

To validate the effectiveness of HoPE, we conducted extensive experiments in various tasks and datasets. We perform "train short, test extended" perplexity evaluations to assess the model's ability to generalize to sequences longer than those seen during training. Additionally, we evaluated our model on long-text benchmarks to test its performance on tasks requiring the processing of extended sequences. The results demonstrate that HoPE outperforms existing positional encoding methods, achieving lower perplexity and better performance on long-text tasks, thereby confirming the superiority of our approach.

The main contributions of this work can be summarized as follows.

\begin{itemize}
\item We revisit classical positional encoding approaches in Transformers and highlight challenges in length extrapolation, distance-based attention biases, and oscillatory position representations.
\item We propose a novel Lorentz rotation framework based on hyperbolic sine and cosine functions, yielding the \emph{Hyperbolic Rotary Positional Encoding} (HoPE) that addresses the oscillatory limitation of RoPE.
\item Through extensive experiments on perplexity metrics and long-sequence benchmarks, we empirically demonstrate the superiority of HoPE in short-to-long generalization and overall positional representation quality.
\end{itemize}

\section{Preliminaries}

\subsection{Relative position encoding}

Let $\mathbb{S}_N = \{w_i\}_{i=1}^N$ be a sequence of $N$ input tokens with $w_i$ being the  $i^{\text{th}}$ element. The corresponding word embedding of $\mathbb{S}_N$ is indicated as $\mathbb{E}_N = \{\boldsymbol{x}_i\}_{i=1}^N$, where $\boldsymbol{x}_i \in \mathbb{R}^d$ is the d-dimensional word embedding vector of token $w_i$ without position information. 

Relative position encoding aims to represent the relative positional relationships between pairs of tokens. In the context of the attention mechanism, this can be expressed as:
\begin{equation}
\small
\langle f_q(x_m, m),\ f_k(x_n, n) \rangle = g(x_m, x_n,\ m - n),
\end{equation}
where $f_q$ and $f_k$ denote the linear transformations in the attention mechanism, and $\langle\cdot, \cdot\rangle$ represents the inner product operation. The core objective of relative positional encoding is to model the relative positional information between tokens at different positions as accurately as possible while satisfying the given formulation.
\\

\subsection{Lorentz Group and Lorentz Transformations}

The Lorentz group is a fundamental concept in theoretical physics, representing the Minkowski spacetime symmetry group in special relativity. It encompasses all linear transformations that preserve the spacetime interval between events, ensuring that the laws of physics remain invariant across different inertial frames. The group is continuous and non-compact, characterized by six parameters corresponding to its generators.

The generators of the Lorentz group can be categorized into two types: rotations and boosts. Rotations correspond to transformations that change the spatial orientation of a reference frame without altering its state of motion. At the same time, boosts relate to changes in the inertial frame's velocity along a particular spatial direction.

\subsubsection{Finite Rotations in Minkowski Space}
Finite rotations around the principal axes in Minkowski spacetime can be represented using specific rotation matrices. These rotations are analogous to those in three-dimensional Euclidean space but are extended to four-dimensional spacetime, preserving the spacetime interval. The rotation matrices for finite rotations about the $x, y$, and $z$ axes are given by:

\begin{equation}
\small
\begin{aligned}
R_x(\theta) &=
\begin{pmatrix}
1 & 0 & 0 & 0 \\
0 & 1 & 0 & 0 \\
0 & 0 & \cos\theta & -\sin\theta \\
0 & 0 & \sin\theta & \cos\theta
\end{pmatrix}, \\
R_y(\psi) &=
\begin{pmatrix}
1 & 0 & 0 & 0 \\
0 & \cos\theta & 0 & \sin\theta \\
0 & 0 & 1 & 0 \\
0 & -\sin\theta & 0 & \cos\theta
\end{pmatrix}, \\
R_z(\theta) &=
\begin{pmatrix}
1 & 0 & 0 & 0 \\
0 & \cos\theta & -\sin\theta & 0 \\
0 & \sin\theta & \cos\theta & 0 \\
0 & 0 & 0 & 1
\end{pmatrix}.
\end{aligned}
\end{equation}

These matrices act on four-dimensional vectors $(c t, x, y, z)^{\top}$ and represent rotations in the spatial components while leaving the temporal component unchanged. Here, $\theta$ is the rotation angles about the $x, y$, and $z$ axes, respectively.

\subsubsection{Lorentz Boosts}

In addition to rotations, the Lorentz group includes boosts, which are transformations between reference frames moving at constant velocities relative to each other. Boosts alter both spatial and temporal components of vectors to preserve the spacetime interval. A standard boost in the $x_{-}$ direction is represented by:

\begin{equation}
\small
B_x(\eta) = \begin{pmatrix}
\cosh \eta & -\sinh \eta & 0 & 0 \\
-\sinh \eta & \cosh \eta & 0 & 0 \\
0 & 0 & 1 & 0 \\
0 & 0 & 0 & 1 \\
\end{pmatrix},
\end{equation}

where $\eta$ is the rapidity parameter, related to the relative velocity $v$ between frames by $\tanh \eta=v / c$ , with $c$ denoting the speed of light. The hyperbolic functions $\cosh \eta$ and $\sinh \eta$ ensure that the spacetime interval remains invariant under the transformation. Similarly, boosts along the $y$ and $z$ axes are represented by corresponding matrices:

\begin{equation}
\small
\begin{aligned}
B_y(\eta) &= \begin{pmatrix}
\cosh \eta & 0 & -\sinh \eta & 0 \\
0 & 1 & 0 & 0 \\
-\sinh \eta & 0 & \cosh \eta & 0 \\
0 & 0 & 0 & 1 \\
\end{pmatrix}, \\
\
B_z(\eta) &= \begin{pmatrix}
\cosh \eta & 0 & 0 & -\sinh \eta \\
0 & 1 & 0 & 0 \\
0 & 0 & 1 & 0 \\
-\sinh \eta & 0 & 0 & \cosh \eta \\
\end{pmatrix}.
\end{aligned}
\end{equation}

\section{HoPE}

In this section, we introduce \textbf{Hyperbolic Rotary Position Encoding (HoPE)}, a positional encoding method formulated from a new perspective. HoPE extends RoPE into hyperbolic space by utilizing Lorentz transformations. This approach leverages the mathematical properties of the Lorentz group to capture complex relative positional relationships within sequences more effectively.

\subsection{Hyperbolic Rotary Position Encoding}

\begin{figure*}[htbp]
  \centering
  \includegraphics[width=0.85\linewidth]{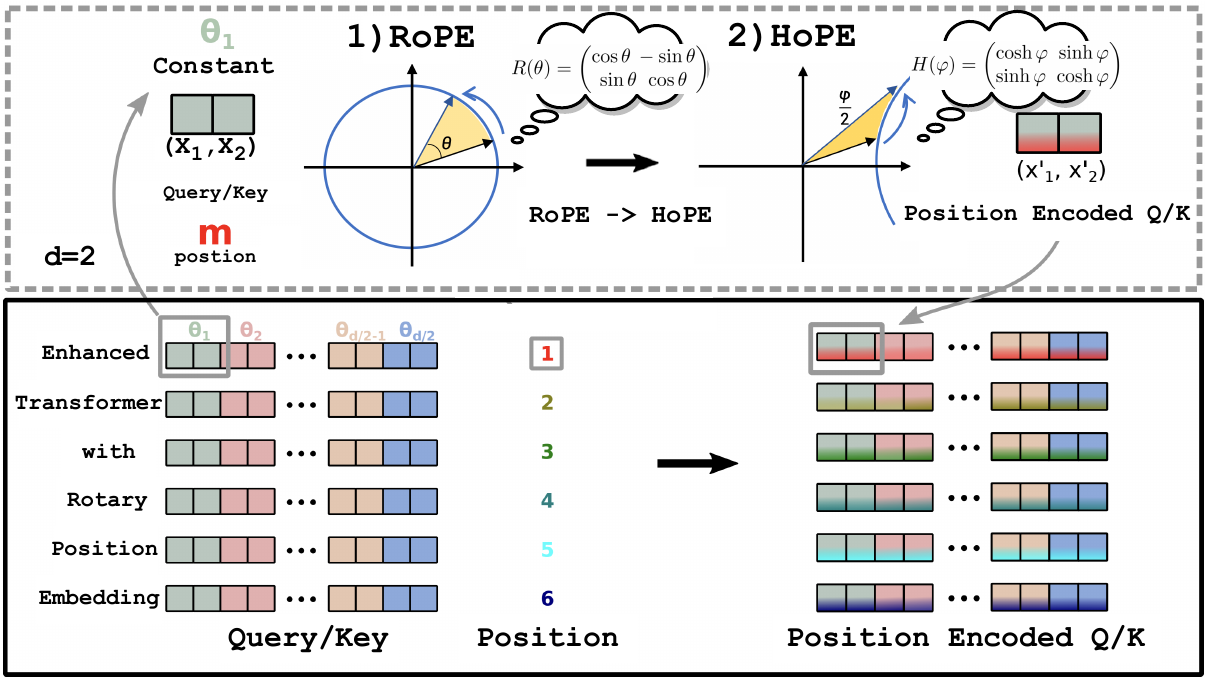} 
  \caption {Implementation of Hyperbolic Rotary Position Embedding.}
\end{figure*}

From the perspective of Lorentz group theory, we draw upon the concept of the \textit{boost generator} in Lorentz transformations to introduce a method of hyperbolic rotation based on hyperbolic trigonometric functions. This approach addresses issues arising from the periodicity inherent in conventional trigonometric functions, such as potential noise due to their cyclic nature.
We define the hyperbolic rotation matrix as follows:

\begin{equation}
B(\theta,m) = 
\begin{pmatrix}
\cosh m\theta & \sinh m\theta  \\
\sinh m\theta & \cosh m\theta  \\
\end{pmatrix}
\end{equation}

Where $\sinh(\cdot)$ and  $\cosh(\cdot)$ represent the hyperbolic sine and cosine functions, respectively. The parameter $\theta$ denotes the rotation angle, while it is a scaling factor that adjusts the transformation's magnitude. 

Like RoPE, we apply rotations of $m\theta$ and $-m\theta$ to the query ($q$) and key ($k$) vectors at position $m$, respectively. Although RoPE may appear to apply identical transformations to both $q$ and $k$, the practical effect, due to the matrix transposition that occurs during the computation of the attention mechanism, results in distinct effective rotations of $m\theta$ for $q$ and $-m\theta$ for $k$. Building upon this concept, we define the rotation matrix for the keys as follows:
\begin{equation}
B'(\theta,m) = 
\begin{pmatrix}
\cosh m\theta & -\sinh m\theta  \\
-\sinh m\theta & \cosh m\theta  \\
\end{pmatrix}
\end{equation}
Based on the above equation, the representations of queries and keys along two specific dimensions in self-attention can be expressed as:
\begin{align}
f_q(\boldsymbol{x}_m, m) &= 
B(\theta,m)
W_{proj}^q
\boldsymbol{x}_m
\end{align}
\begin{align}
f_k(\boldsymbol{x}_m, m) &= 
B'(\theta,m)
W_{proj}^k
\boldsymbol{x}_m
\end{align}
Here, $W_{proj}^q$ and $W_{proj}^k$ represent the projection layer weights for queries and keys, respectively, and $\boldsymbol{x}_m$ denotes the original token embedding at position $m$. This formalism ensures that the learned positional representations are distinctly handled for queries and keys, leveraging the unique properties of rotational transformations in capturing positional information.

As a result of the action of the attention mechanism, the final form of the dot product is obtained as follows:
\begin{multline}
g(\boldsymbol{x}_m, \boldsymbol{x}_n, n - m) = 
\langle f_q(\boldsymbol{x}_m, m), f_k(\boldsymbol{x}_n, n) \rangle\\
= e^{-(m-n)\theta^{'}}
\begin{pmatrix}
q_m^{(1)} & q_m^{(2)}
\end{pmatrix}
B(\theta,m-n)
\begin{pmatrix}
k_n^{(1)} \\
k_n^{(2)}
\end{pmatrix}
\end{multline}

However, we find that such transformations alone do not satisfy the assumptions of positional encoding. This is because, unlike RoPE, the Boost generator is not an orthogonal matrix. Furthermore, due to the monotonicity of hyperbolic trigonometric functions, the difference $m-n$ increases the dot product of $q$ and $k$. As $m-n$ increases, the calculated attention weight of $q$ and $k$ also increases. This contradicts the assumption of positional encoding, which posits that tokens closer to each other should be assigned higher attention weights.

To address this issue, we introduce a penalty coefficient $e^{\pm m \theta'}$, where $\theta'$ is a learnable or predefined parameter, to modulate the positional impact on the dot product. Specifically, the penalty ensures that as the positional difference $m-n$ increases, the dot product of $q$ and $k$ decreases, thereby enforcing the intended behaviour of the attention mechanism. This design ensures that tokens closer in position are prioritized with higher attention weights, enhancing the model's ability to capture local context and long-range dependencies accurately.

The modified query and key representations, which incorporate this penalty coefficient, are defined as follows:
\begin{equation}
f_q(\boldsymbol{x}_m, m) = e^{-m\theta'}
B(\theta, m)
W_{\text{proj}}^q
\boldsymbol{x}_m,
\end{equation}
\begin{equation}
f_k(\boldsymbol{x}_m, m) = e^{m\theta'}
B'(\theta, m)
W_{\text{proj}}^k
\boldsymbol{x}_m,
\end{equation}
 
 Using $e^{\pm m \theta'}$ ensures a decaying or amplifying effect on the positional encoding component, countering the undesired monotonic increase and aligning the attention mechanism with the theoretical assumptions of positional encodings.

\subsection{Theoretical Analysis}

\subsubsection{Long-range Decay Property}
For dimension-pair $(2 i, 2 i+1)$, consider the asymptotic behavior:
\begin{equation}
\small
\lim_{|m-n|\to\infty} e^{-|m-n|\theta'}\cosh(|m-n|\theta_i) \propto e^{-|m-n|(\theta'-\theta_i)}
\end{equation}

When $\theta^{\prime}>\theta_i,\forall i$, the attention weights exhibit exponential decay concerning positional distance. This satisfies the locality prior while maintaining controlled long-range interaction capability.

\subsubsection{Positional Information Capacity}
HoPE preserves RoPE's theoretical advantages in positional discrimination:

\noindent \textbf{Positional Discrimination Capacity}. For any relative position $r \in \mathbb{Z}$ and query vector $q, ~ \exists$ key vector $k$ such that:
\begin{equation}
\underset{s}{\mathrm{argmax}}(\langle f_q(m), f_k(m+s) \rangle) = r
\end{equation}\label{thm:1}
\noindent The hyperbolic rotation creates position-dependent orientation in the embedding space. For target position $r$, construct $k$ such that $W_k \boldsymbol{x}_n$ aligns with $e^{r \theta^{\prime}} R^{\prime}(-\theta, r) W_q \boldsymbol{x}_m$ in the rotated space. This attribute ensures that HoPE maintains the capability to focus on tokens at significant positions while introducing a controllable distance decay.

\subsubsection{Generalization to Higher Dimensions}
For $d$-dimensional embeddings ( $d$ even), we implement block-diagonal transformations:
\begin{equation}
R^d_{\Theta,m} = \bigoplus_{i=0}^{d/2-1} R(\theta_i,m)
\end{equation}
Each 2D subspace receives independent rotation parameters $\theta_i$, with global damping controlled by $\theta^{\prime}>\max _i \theta_i$. The parameterization enables hierarchical capture of positional relationships.
These refinements maintain the original contributions while improving mathematical rigor, clarifying causal relationships between components, and emphasizing the approach's theoretical underpinnings. The narrative flows better, connecting motivation, implementation, and theoretical analysis.

\section{Experiments}
\subsection{Experimental Setup}
We evaluate the HoPE's positional encoding capability using two primary metrics: perplexity for pre-training and performance on downstream tasks with the SCROLLS benchmark.

\textbf{Perplexity}: This metric quantifies the ability of a language model to predict a sequence of words or tokens, with lower perplexity indicating greater confidence and accuracy in its predictions. 

\textbf{Downstream Task Performance}: In natural language processing tasks, downstream performance does not always correlate directly with perplexity. Consequently, we have chosen to use the SCROLLS\cite{shaham2022scrollsstandardizedcomparisonlong} benchmark to evaluate the impact of various positional encodings on the performance of the downstream task.

\subsection{Perplexity Experiment (PPL)}

\begin{table*}[h!]
    \centering
    \renewcommand\arraystretch{1.1}
    \setlength{\tabcolsep}{12pt}
    \begin{tabular}{lcccccc} 
        \hline
        \textbf{Method} & \textbf{1024} & \textbf{2048} & \textbf{3072} & \textbf{4096} & \textbf{5120} & \textbf{6144} \\
        \hline
        RoPE            & 12.82         & 25.80         & 56.28         & 88.59         & 116.63        & 144.13        \\
        Alibi           & \textbf{11.95}& 25.11         & 52.54         & 79.04         & 107.59        & 132.80        \\
        HoPE            & 13.35         & \textbf{16.46}& \textbf{35.07}& \textbf{60.03}& \textbf{85.94}& \textbf{110.02}\\
        \hline
        Bipe-RoPE       & 13.74         & 14.49         & 25.05         & 40.50         & 54.47         & 66.64         \\
        Bipe-Alibi      & \textbf{11.95}& 29.06         & 63.19         & 91.86         & 118.54        & 142.05        \\
        Bipe-HoPE       & 13.71         & \textbf{14.47}& \textbf{24.00}& \textbf{38.78}& \textbf{52.84}& \textbf{65.34}\\
        \hline
    \end{tabular}
    \caption{Perplexity Performance Comparison on PG19}
    \label{tab:1}
\end{table*}

\begin{table*}[h!]
    \centering
    \renewcommand\arraystretch{1.1}
    \setlength{\tabcolsep}{12pt}
    \begin{tabular}{lcccccc} 
        \hline

        \textbf{Method} & \textbf{1024} & \textbf{2048} & \textbf{3072} & \textbf{4096} & \textbf{5120} & \textbf{6144} \\
         \hline
        RoPE            &4.81&12.11&36.84 &62.78&98.21&132.63        \\
        Alibi           &4.82&\textbf{4.89}&\textbf{4.93}&\textbf{4.98}&\textbf{4.95}&\textbf{5.01}        \\
        HoPE            &\textbf{4.78}&8.90&22.28&40.10&60.48&82.04\\
        \hline
        Bipe-RoPE       &\textbf{4.74}&5.73&12.84&20.05&30.84&40.58         \\
        Bipe-Alibi      &4.82&\textbf{4.88}&\textbf{4.94}&\textbf{4.97}&\textbf{4.92}&\textbf{4.98}        \\
        Bipe-HoPE       &4.83&5.12&12.20&20.01&28.85&38.34\\
        \hline
    \end{tabular}
    \caption{Perplexity Performance Comparison on arXiv}
    \label{tab:2}
\end{table*}

We test the length extrapolation capability of Transformer-based language models with various positional encoding methods. Following the methodology of ~\cite{chi2022kerplekernelizedrelativepositional}, we use the Pile dataset \cite{gao2020pile800gbdatasetdiverse}as the pre-training corpus and evaluate the log perplexity of pre-trained language models in the test sets of PG19 \cite{rae2019compressivetransformerslongrangesequence} and arXiv. We conduct non-overlapping evaluations when computing the perplexity score.

The pre-training sequence length is set to 1024, and we evaluate zero-shot perplexity on sequence lengths [1024, 2048, 3072, 4096, 5120, 6144]. We choose the standard decoder-only Transformer\cite{touvron2023llama2openfoundation} as the base model and compare our HoPE method against other positional encoding methods: RoPE and Alibi  For general segmentation purposes, and full stops determine the boundaries ""."" and newline characters "\verb|\n|" The Transformer-based language model configuration includes 12 layers, a hidden dimension of 768, and 12 attention heads, resulting in approximately 155M parameters.

The results, illustrated in Table~\ref{tab:1} and Table~\ref{tab:2}(Best performing results
are highlighted in bold), show that our HoPE method consistently outperforms RoPE on sequences longer than the training length. While Alibi achieves the lowest perplexity on the arXiv dataset, as noted in \cite{chen2024clexcontinuouslengthextrapolation,peng2023yarnefficientcontextwindow}, this phenomenon can be attributed to two main factors:

1) The nature of the training corpus: Most texts in our pre-training datasets predominantly exhibit short-distance dependencies, meaning that accurate token prediction primarily relies on information from nearby contexts rather than long-range dependencies.

2) Alibi's architectural advantage in this scenario: Its linear attention decay mechanism naturally emphasizes local context while attenuating long-distance information. This characteristic aligns well with the short-distance dependency pattern in our training data, resulting in stable perplexity scores even as sequence length increases.

\noindent \textbf{Integrating Hope with interpolation strategies}.
Recent advancements in the extrapolation of context length involve interpolating language models based on relative positional encoding, utilizing segmented position encoding techniques \cite {golovneva2024contextualpositionencodinglearning}. To investigate further the performance differences between our HoPE method and other relative positional encodings post-fine-tuning, we employ BiPE \cite{he2024stoneshitbirdbilevel} to fine-tune language models pre-trained on the PG19 and arXiv datasets. We then evaluated their performance on downstream tasks. The results are presented in Table~\ref{tab:1} and Table~\ref{tab:2}. Similarly to the observations before fine-tuning, BiPE-HoPE outperforms BiPE-RoPE on longer sequences after fine-tuning.

\subsection{Fine-Tuning Experiment}

To evaluate the model's performance in understanding extended contexts, following \cite{ainslie2023colt5fasterlongrangetransformers}, we further fine-tune the pre-trained checkpoints on the SCROLLS benchmark  SCROLLS consists of seven distinct datasets covering various tasks  We employ three evaluation metrics for different tasks: RGL score (ROUGE-L), unigram overlap (F1), and exact match (EM)  We fine-tune pre-trained models using a sequence length of 8192 and select the last model checkpoint on the validation set for final evaluation.

As shown in Table~\ref{tab:3}, although Alibi achieved the best extrapolation performance on the PPL task using the arXiv dataset, HOPE emerged as the superior positional encoding when fine-tuning with a sequence length of 8192 for downstream tasks. Specifically, HOPE outperformed other positional encodings in four out of seven tasks. In the NarrativeQA task, HOPE scored 1.57 points higher in Rouge-L compared to RoPE, and in the QMSum task, it scored 2.33 points higher than Alibi.
\begin{table*}[h!]
    \centering
    \setlength{\tabcolsep}{10pt}
    \renewcommand\arraystretch{1.1}
    \begin{tabular}{lccccccc} 
        \hline
         & \textbf{QAS} & \textbf{CNLI} & \textbf{QMS} & \textbf{NQA} & \textbf{SumS} & \textbf{GovR}&\textbf{QuAL}\\
        \hline
        \textbf{Metric}&F1&EM&RGL&F1&RGL&RGL&EM\\
        \hline
        \textbf{Median length}&5472&2148&14197&57829&9046&8841&7171\\
        \hline
        Sinusoidal&9.2&58.9&16.89&6.08&12.74&14.47&1.9\\
        Randomized RoPE&13.02&69.37&16.31&6.89&13.45&16.94&\textbf{12.8}\\
        \hline
        RoPE&12.98&\textbf{69.43}&16.03&7.57&\textbf{13.69} &15.55&0.68\\
        Alibi&14.17&65.41&14.75&6.73&13.04&18.83&0.87\\
        HoPE&\textbf{14.52}&68.91&\textbf{17.08}&\textbf{9.14}&13.35&\textbf{19.34}& 0.45\\
        \hline
    \end{tabular}
    \caption{Performance comparison on SCROLLS benchmark. Abbreviations for dataset names: Qasper (Qas), ContractNLI (CNLI),
QMSum (QMS), NarrativeQA (NQA), SummScreenFD (SumS), GovReport (GovR), and QuALITY (QuAL).Best performing results
are highlighted in \textbf{bold}.}
    \label{tab:3}
\end{table*}
\subsection{Ablation Studies}

To further analyze HoPE's effectiveness, we conduct ablation studies that examine the impact of individual components. These studies provide deeper insight into the aspects of HoPE that contribute significantly to its overall performance. The relevant sections of this paper provide Detailed results and discussions of these ablation studies.

\noindent \textbf{Scaling factor is important}: We investigated the impact of scaling factors on overall positional encoding within perplexity experiments conducted on arXiv. Specifically, we fixed the hyperbolic rotation angle and observed the effects of different scaling factors on the perplexity of the model.

In these experiments, the scaling factor was defined as the proportion used to adjust the magnitude of positional encodings. Our experimental results indicate that different scaling factors significantly affect the model's perplexity with the rotation angle held constant. As shown in Figure \ref{fig:ablation}:

\begin{figure}[ht]
    \centering 
    \includegraphics[width=0.51\textwidth]{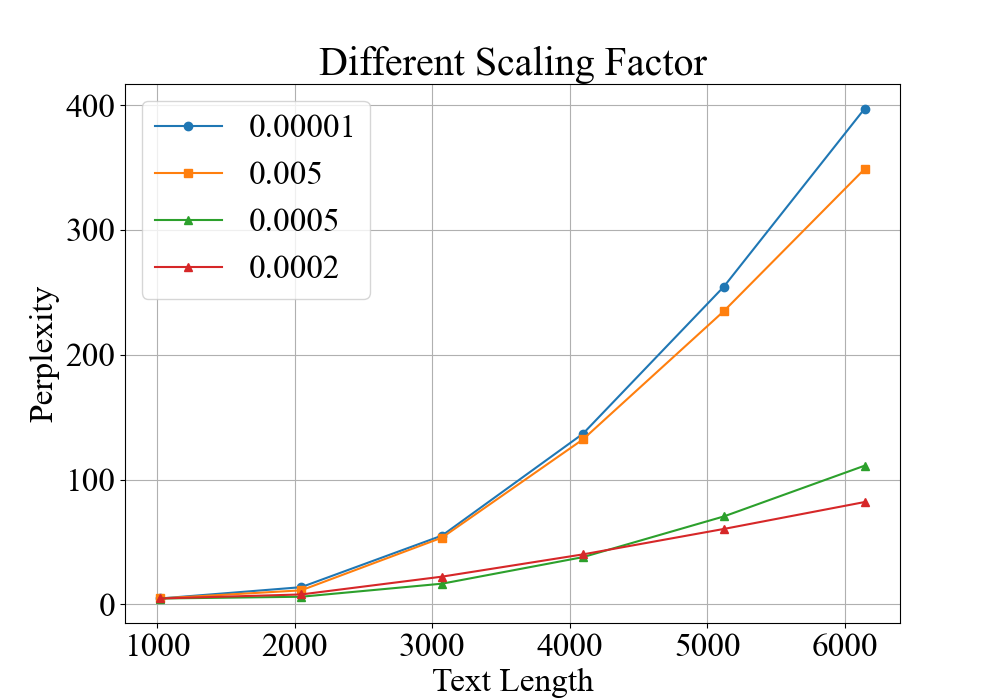} 
    \caption{Ablation Experiment} 
    \label{fig:ablation} 
\end{figure}

\noindent \textbf{Smaller Scaling Factors}: When the scaling factor is small, the amplitude of positional encodings decreases, making it difficult for the model to capture long-range dependencies within sequences, resulting in higher perplexity.

\noindent \textbf{Moderate Scaling Factors}: Moderate scaling factors strike a balance by maintaining positional information while avoiding noise amplification or unnecessary details, typically leading to lower perplexity.

\noindent \textbf{Larger Scaling Factors}: Substantial scaling factors can amplify noise or other non-ideal characteristics in positional encodings, deteriorating model performance and resulting in higher perplexity.

\subsection{Further Analysis from the Attention Weight Perspective}
\begin{figure}[ht]
    \centering 
    \includegraphics[width=0.48\textwidth]{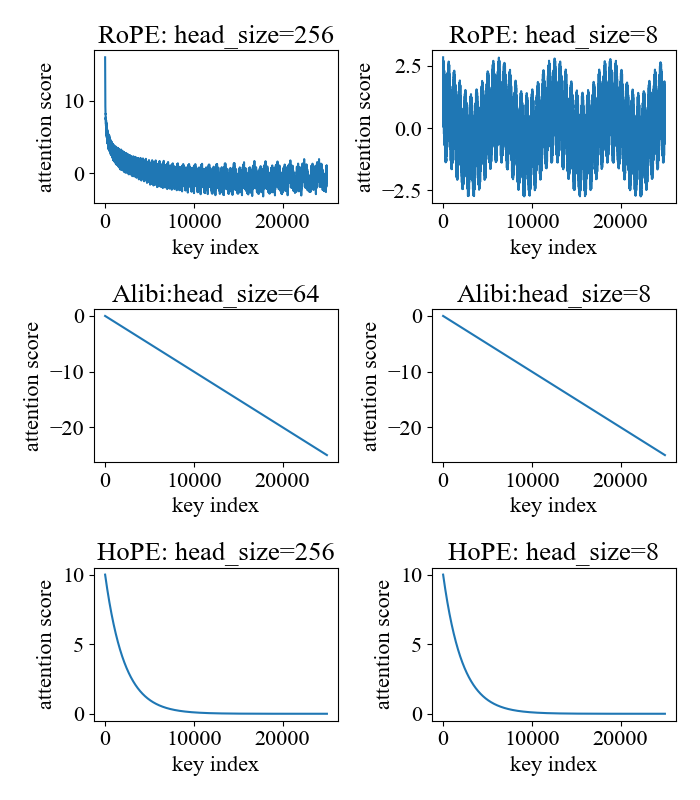} 
    \caption{Attention Weight Values under Different Positional Encodings} 
    \label{fig:attention} 
\end{figure}

To further analyze the capabilities of various positional encodings, we investigated their impact on attention weights before activation functions are applied. Specifically, we initialized two fixed vectors, q and k, to simulate the effect of different positional encodings on computing attention weights when these vectors are placed at varying positions. The results are illustrated in Figure \ref{fig:attention}.
All positional encodings exhibit a decay characteristic with increasing relative distance. However, during this decay process, Rope demonstrates fluctuation issues due to the periodic nature of trigonometric functions, leading to localized decreases in higher frequency dimensions. In contrast, the decline observed in Alibi is not smooth, suggesting that it could be more accurately described as an attention bias rather than a positional encoding. In particular, the Hope was meticulously designed to overcome these challenges, resulting in smoother and more stable performance across different relative distances.
\\
\begin{figure}[ht]
    \centering 
    \includegraphics[width=0.49\textwidth]{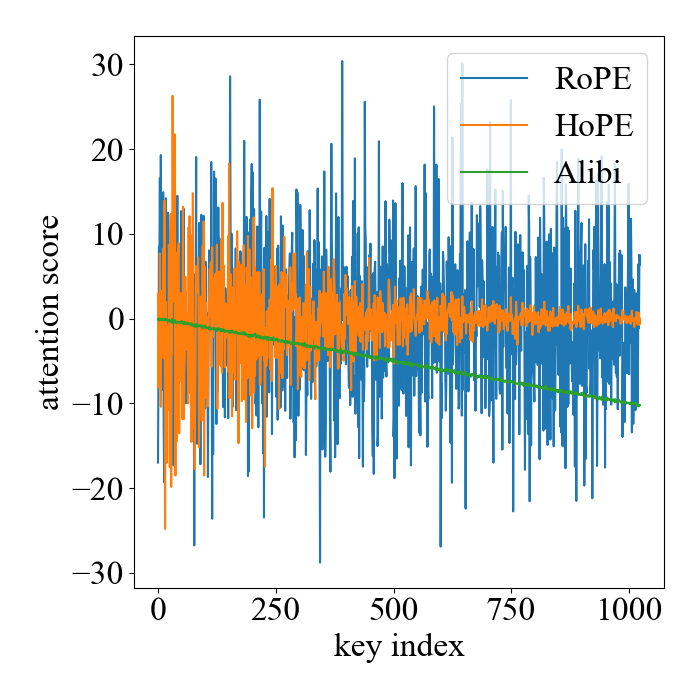} 
    \caption{Attention Weight Values under Different Positional Encodings} 
    \label{fig:gauss} 
\end{figure}
Furthermore, instead of assigning fixed values to these vectors, each q and k vector was independently initialized using a Gaussian distribution, which was more reflective of real-world scenarios. The resulting attention weights, derived from the inner product of q and k, were plotted in Fig. \ref{fig:gauss}. Our findings reveal that RoPE modifies the inner product between q and k vectors by adjusting their angular relationships  Although RoPE effectively captures long-range dependencies when q and k vectors are aligned in a fixed direction, this alignment does not consistently apply across all dimensions of high-dimensional vectors when q and k are randomly initialized  Specifically, RoPE's ability to preserve locality among closely related dimensions appears limited; random initialization of q and k (implying random angles between them) diminishes RoPE's capacity to maintain local relevance, resembling situations without positional encoding (NoPE)  Furthermore, initializing vectors with a Gaussian distribution suggests that specific tokens should attract more attention than others, with attention decreasing as distance increases a behavior consistent with our expectations  However, one drawback observed with AliBi is that distant but relevant tokens sometimes receive less attention than closer, irrelevant ones.

\section{Related work}

Existing positional encoding mechanisms can be classified into absolute and relative positional encodings, each having different trade-offs in length extrapolation and dependency modeling. Absolute Positional Encodings, pioneered by \cite{vaswani2023attentionneed} using fixed sinusoidal patterns or learnable embeddings, are simple but struggle with length extrapolation due to their reliance on predefined positional indices \cite{devlin2019bertpretrainingdeepbidirectional}. Studies \cite{wang2020positionembeddingslearnempirical} have shown that while sinusoidal embeddings can implicitly capture positional relationships, they fail to generalize effectively beyond training sequence lengths. Relative Positional Encodings, introduced by \cite{shaw2018selfattentionrelativepositionrepresentations}, model pairwise token distances through additive biases in attention scores, offering greater flexibility but encountering scalability challenges with long-range dependencies. Alibi \cite{press2022trainshorttestlong} enhanced extrapolation capability through a distance-decaying linear attention bias, but its heuristic design lacks guarantees for monotonic decay, leading to suboptimal performance on extremely long sequences. Rotary Positional Encodings (RoPE) \cite{su2023roformerenhancedtransformerrotary}, implementing rotation-based positional encoding through trigonometric transformations, theoretically maintains relative positional relationships across different sequence lengths. However, RoPE exhibits oscillatory attention patterns due to its trigonometric periodicity, which can destabilize long-distance dependency modeling \cite{barbero2024roundroundgomakes}.

Hyperbolic space, regarded as the continuous analogue of discrete trees~\cite{2010hyperbolic}, offers inherent advantages for modeling data with implicit or explicit tree-like structures, including hierarchical organizations and power-law distributions~\cite{adcock2013tree,zhou2022telegraph}. Recent advances in representation learning have extensively demonstrated the superiority of hyperbolic geometry through multiple perspectives: low-distortion embeddings~\cite{sarkar2011low}, reduced generalization error~\cite{suzuki2021generalization1,suzuki2021generalization2}, and superior empirical performance~\cite{hgcn2019,yang2022hicf}. These advantages have been successfully leveraged across diverse research domains and downstream applications~\cite{peng2021hyperbolic,hyper5,hyper1,hyper2,mettes2023hyperbolic,hyper3,hyper4,yang2021hyper}, encompassing graph learning, computer vision, and natural language processing.
\section{Conclusion}
\label{sec:conclusion }
We propose HoPE to address the oscillatory attention patterns in RoPE that limit long-range dependency modeling. Based on Lorentz transformations, HoPE replaces trigonometric rotations with hyperbolic functions to ensure monotonic attention decay with increasing token distances. Theoretical analysis validates that HoPE's geometric design naturally achieves distance-aware attention.
Extensive experiments demonstrate HoPE's advantages over existing methods. It exhibits superior length extrapolation capabilities in "train short, test long" scenarios and achieves state-of-the-art performance on long-text tasks. These results verify that HoPE effectively maintains stable positional representations while preserving the essential rotational invariance of Transformer attention.

\section{Limitations}
In this paper, we propose HoPE to address limitations in current positional encoding methods. By leveraging Lorentz transformations and hyperbolic functions, HoPE yields more stable position representations. Theoretical analysis and extensive experiments demonstrate its advantages in capturing long-range dependencies and enabling length extrapolation. However, HoPE still faces challenges. First, while it excels in text-only tasks, its performance in multimodal scenarios (where text, audio, and visual inputs must be jointly modelled) remains unverified. Second, the method's effectiveness hinges on careful tuning of the damping coefficient $\theta'$. Suboptimal choices can degrade performance, especially for tasks with varying positional sensitivity requirements.

\bibliography{custom}

\begin{thebibliography}{53}
\providecommand{\natexlab}[1]{#1}

\bibitem[{Adcock et~al.(2013)Adcock, Sullivan, and Mahoney}]{adcock2013tree}
Aaron~B Adcock, Blair~D Sullivan, and Michael~W Mahoney. 2013.
\newblock Tree-like structure in large social and information networks.
\newblock In \emph{IEEE International Conference on Data Mining}, pages 1--10.
  IEEE.

\bibitem[{Ainslie et~al.(2023)Ainslie, Lei, de~Jong, Ontañón, Brahma,
  Zemlyanskiy, Uthus, Guo, Lee-Thorp, Tay, Sung, and
  Sanghai}]{ainslie2023colt5fasterlongrangetransformers}
Joshua Ainslie, Tao Lei, Michiel de~Jong, Santiago Ontañón, Siddhartha
  Brahma, Yury Zemlyanskiy, David Uthus, Mandy Guo, James Lee-Thorp, Yi~Tay,
  Yun-Hsuan Sung, and Sumit Sanghai. 2023.
\newblock \href {https://arxiv.org/abs/2303.09752} {Colt5: Faster long-range
  transformers with conditional computation}.
\newblock \emph{Preprint}, arXiv:2303.09752.

\bibitem[{Anil et~al.(2022)Anil, Wu, Andreassen, Lewkowycz, Misra, Ramasesh,
  Slone, Gur-Ari, Dyer, and
  Neyshabur}]{anil2022exploringlengthgeneralizationlarge}
Cem Anil, Yuhuai Wu, Anders Andreassen, Aitor Lewkowycz, Vedant Misra, Vinay
  Ramasesh, Ambrose Slone, Guy Gur-Ari, Ethan Dyer, and Behnam Neyshabur. 2022.
\newblock \href {https://arxiv.org/abs/2207.04901} {Exploring length
  generalization in large language models}.
\newblock \emph{Preprint}, arXiv:2207.04901.

\bibitem[{Barbero et~al.(2024)Barbero, Vitvitskyi, Perivolaropoulos, Pascanu,
  and Veličković}]{barbero2024roundroundgomakes}
Federico Barbero, Alex Vitvitskyi, Christos Perivolaropoulos, Razvan Pascanu,
  and Petar Veličković. 2024.
\newblock \href {https://arxiv.org/abs/2410.06205} {Round and round we go! what
  makes rotary positional encodings useful?}
\newblock \emph{Preprint}, arXiv:2410.06205.

\bibitem[{Chami et~al.(2019)Chami, Ying, R{\'e}, and Leskovec}]{hgcn2019}
Ines Chami, Zhitao Ying, Christopher R{\'e}, and Jure Leskovec. 2019.
\newblock Hyperbolic graph convolutional neural networks.
\newblock In \emph{Advances in Neural Information Processing Systems}, pages
  4868--4879.

\bibitem[{Chen et~al.(2024)Chen, Li, Meng, Liang, and
  Bing}]{chen2024clexcontinuouslengthextrapolation}
Guanzheng Chen, Xin Li, Zaiqiao Meng, Shangsong Liang, and Lidong Bing. 2024.
\newblock \href {https://arxiv.org/abs/2310.16450} {Clex: Continuous length
  extrapolation for large language models}.
\newblock \emph{Preprint}, arXiv:2310.16450.

\bibitem[{Chen et~al.(2022)Chen, Chu, Wiseman, and
  Gimpel}]{chen2022summscreendatasetabstractivescreenplay}
Mingda Chen, Zewei Chu, Sam Wiseman, and Kevin Gimpel. 2022.
\newblock \href {https://arxiv.org/abs/2104.07091} {Summscreen: A dataset for
  abstractive screenplay summarization}.
\newblock \emph{Preprint}, arXiv:2104.07091.

\bibitem[{Chen et~al.(2023)Chen, Wong, Chen, and
  Tian}]{chen2023extendingcontextwindowlarge}
Shouyuan Chen, Sherman Wong, Liangjian Chen, and Yuandong Tian. 2023.
\newblock \href {https://arxiv.org/abs/2306.15595} {Extending context window of
  large language models via positional interpolation}.
\newblock \emph{Preprint}, arXiv:2306.15595.

\bibitem[{Chi et~al.(2022{\natexlab{a}})Chi, Fan, Ramadge, and
  Rudnicky}]{chi2022kerple}
Ta-Chung Chi, Ting-Han Fan, Peter~J Ramadge, and Alexander Rudnicky.
  2022{\natexlab{a}}.
\newblock Kerple: Kernelized relative positional embedding for length
  extrapolation.
\newblock \emph{Advances in Neural Information Processing Systems},
  35:8386--8399.

\bibitem[{Chi et~al.(2022{\natexlab{b}})Chi, Fan, Ramadge, and
  Rudnicky}]{chi2022kerplekernelizedrelativepositional}
Ta-Chung Chi, Ting-Han Fan, Peter~J. Ramadge, and Alexander~I. Rudnicky.
  2022{\natexlab{b}}.
\newblock \href {https://arxiv.org/abs/2205.09921} {Kerple: Kernelized relative
  positional embedding for length extrapolation}.
\newblock \emph{Preprint}, arXiv:2205.09921.

\bibitem[{Chi et~al.(2023)Chi, Fan, Ramadge et~al.}]{chi2023dissecting}
Ta-Chung Chi, Ting-Han Fan, Peter~J Ramadge, et~al. 2023.
\newblock Dissecting transformer length extrapolation via the lens of receptive
  field analysis.
\newblock In \emph{The 61st Annual Meeting Of The Association For Computational
  Linguistics}.

\bibitem[{Chowdhury and
  Caragea(2023)}]{chowdhury2023monotoniclocationattentionlength}
Jishnu~Ray Chowdhury and Cornelia Caragea. 2023.
\newblock \href {https://arxiv.org/abs/2305.20019} {Monotonic location
  attention for length generalization}.
\newblock \emph{Preprint}, arXiv:2305.20019.

\bibitem[{Dasigi et~al.(2021)Dasigi, Lo, Beltagy, Cohan, Smith, and
  Gardner}]{dasigi2021datasetinformationseekingquestionsanswers}
Pradeep Dasigi, Kyle Lo, Iz~Beltagy, Arman Cohan, Noah~A. Smith, and Matt
  Gardner. 2021.
\newblock \href {https://arxiv.org/abs/2105.03011} {A dataset of
  information-seeking questions and answers anchored in research papers}.
\newblock \emph{Preprint}, arXiv:2105.03011.

\bibitem[{Devlin et~al.(2019)Devlin, Chang, Lee, and
  Toutanova}]{devlin2019bertpretrainingdeepbidirectional}
Jacob Devlin, Ming-Wei Chang, Kenton Lee, and Kristina Toutanova. 2019.
\newblock \href {https://arxiv.org/abs/1810.04805} {Bert: Pre-training of deep
  bidirectional transformers for language understanding}.
\newblock \emph{Preprint}, arXiv:1810.04805.

\bibitem[{Gao et~al.(2020)Gao, Biderman, Black, Golding, Hoppe, Foster, Phang,
  He, Thite, Nabeshima, Presser, and Leahy}]{gao2020pile800gbdatasetdiverse}
Leo Gao, Stella Biderman, Sid Black, Laurence Golding, Travis Hoppe, Charles
  Foster, Jason Phang, Horace He, Anish Thite, Noa Nabeshima, Shawn Presser,
  and Connor Leahy. 2020.
\newblock \href {https://arxiv.org/abs/2101.00027} {The pile: An 800gb dataset
  of diverse text for language modeling}.
\newblock \emph{Preprint}, arXiv:2101.00027.

\bibitem[{Golovneva et~al.(2024)Golovneva, Wang, Weston, and
  Sukhbaatar}]{golovneva2024contextualpositionencodinglearning}
Olga Golovneva, Tianlu Wang, Jason Weston, and Sainbayar Sukhbaatar. 2024.
\newblock \href {https://arxiv.org/abs/2405.18719} {Contextual position
  encoding: Learning to count what's important}.
\newblock \emph{Preprint}, arXiv:2405.18719.

\bibitem[{Hall(2000)}]{hall2000elementaryintroductiongroupsrepresentations}
Brian~C. Hall. 2000.
\newblock \href {https://arxiv.org/abs/math-ph/0005032} {An elementary
  introduction to groups and representations}.
\newblock \emph{Preprint}, arXiv:math-ph/0005032.

\bibitem[{Haviv et~al.(2022)Haviv, Ram, Press, Izsak, and
  Levy}]{haviv2022transformerlanguagemodelspositional}
Adi Haviv, Ori Ram, Ofir Press, Peter Izsak, and Omer Levy. 2022.
\newblock \href {https://arxiv.org/abs/2203.16634} {Transformer language models
  without positional encodings still learn positional information}.
\newblock \emph{Preprint}, arXiv:2203.16634.

\bibitem[{He et~al.(2024)He, Feng, Luo, Yang, Wang, Xu, Zhang, Yang, and
  He}]{he2024stoneshitbirdbilevel}
Zhenyu He, Guhao Feng, Shengjie Luo, Kai Yang, Liwei Wang, Jingjing Xu, Zhi
  Zhang, Hongxia Yang, and Di~He. 2024.
\newblock \href {https://arxiv.org/abs/2401.16421} {Two stones hit one bird:
  Bilevel positional encoding for better length extrapolation}.
\newblock \emph{Preprint}, arXiv:2401.16421.

\bibitem[{Huang et~al.(2021)Huang, Cao, Parulian, Ji, and
  Wang}]{huang2021efficientattentionslongdocument}
Luyang Huang, Shuyang Cao, Nikolaus Parulian, Heng Ji, and Lu~Wang. 2021.
\newblock \href {https://arxiv.org/abs/2104.02112} {Efficient attentions for
  long document summarization}.
\newblock \emph{Preprint}, arXiv:2104.02112.

\bibitem[{Kazemnejad et~al.(2023)Kazemnejad, Padhi, Ramamurthy, Das, and
  Reddy}]{kazemnejad2023impactpositionalencodinglength}
Amirhossein Kazemnejad, Inkit Padhi, Karthikeyan~Natesan Ramamurthy, Payel Das,
  and Siva Reddy. 2023.
\newblock \href {https://arxiv.org/abs/2305.19466} {The impact of positional
  encoding on length generalization in transformers}.
\newblock \emph{Preprint}, arXiv:2305.19466.

\bibitem[{Koreeda and
  Manning(2021)}]{koreeda2021contractnlidatasetdocumentlevelnatural}
Yuta Koreeda and Christopher~D. Manning. 2021.
\newblock \href {https://arxiv.org/abs/2110.01799} {Contractnli: A dataset for
  document-level natural language inference for contracts}.
\newblock \emph{Preprint}, arXiv:2110.01799.

\bibitem[{Kočiský et~al.(2017)Kočiský, Schwarz, Blunsom, Dyer, Hermann,
  Melis, and
  Grefenstette}]{kočiský2017narrativeqareadingcomprehensionchallenge}
Tomáš Kočiský, Jonathan Schwarz, Phil Blunsom, Chris Dyer, Karl~Moritz
  Hermann, Gábor Melis, and Edward Grefenstette. 2017.
\newblock \href {https://arxiv.org/abs/1712.07040} {The narrativeqa reading
  comprehension challenge}.
\newblock \emph{Preprint}, arXiv:1712.07040.

\bibitem[{Krioukov et~al.(2010)Krioukov, Papadopoulos, Kitsak, Vahdat, and
  Bogun{\'a}}]{2010hyperbolic}
Dmitri Krioukov, Fragkiskos Papadopoulos, Maksim Kitsak, Amin Vahdat, and
  Mari{\'a}n Bogun{\'a}. 2010.
\newblock Hyperbolic geometry of complex networks.
\newblock \emph{Physical Review E}, 82(3):036106.

\bibitem[{Li et~al.(2024)Li, You, Guruganesh, Ainslie, Ontanon, Zaheer,
  Sanghai, Yang, Kumar, and
  Bhojanapalli}]{li2024functionalinterpolationrelativepositions}
Shanda Li, Chong You, Guru Guruganesh, Joshua Ainslie, Santiago Ontanon, Manzil
  Zaheer, Sumit Sanghai, Yiming Yang, Sanjiv Kumar, and Srinadh Bhojanapalli.
  2024.
\newblock \href {https://arxiv.org/abs/2310.04418} {Functional interpolation
  for relative positions improves long context transformers}.
\newblock \emph{Preprint}, arXiv:2310.04418.

\bibitem[{Mettes et~al.(2023)Mettes, Atigh, Keller-Ressel, Gu, and
  Yeung}]{mettes2023hyperbolic}
Pascal Mettes, Mina~Ghadimi Atigh, Martin Keller-Ressel, Jeffrey Gu, and Serena
  Yeung. 2023.
\newblock Hyperbolic deep learning in computer vision: A survey.
\newblock \emph{arXiv preprint arXiv:2305.06611}.

\bibitem[{Pang et~al.(2022)Pang, Parrish, Joshi, Nangia, Phang, Chen,
  Padmakumar, Ma, Thompson, He, and
  Bowman}]{pang2022qualityquestionansweringlong}
Richard~Yuanzhe Pang, Alicia Parrish, Nitish Joshi, Nikita Nangia, Jason Phang,
  Angelica Chen, Vishakh Padmakumar, Johnny Ma, Jana Thompson, He~He, and
  Samuel~R. Bowman. 2022.
\newblock \href {https://arxiv.org/abs/2112.08608} {Quality: Question answering
  with long input texts, yes!}
\newblock \emph{Preprint}, arXiv:2112.08608.

\bibitem[{Peng et~al.(2023)Peng, Quesnelle, Fan, and
  Shippole}]{peng2023yarnefficientcontextwindow}
Bowen Peng, Jeffrey Quesnelle, Honglu Fan, and Enrico Shippole. 2023.
\newblock \href {https://arxiv.org/abs/2309.00071} {Yarn: Efficient context
  window extension of large language models}.
\newblock \emph{Preprint}, arXiv:2309.00071.

\bibitem[{Peng et~al.(2021)Peng, Varanka, Mostafa, Shi, and
  Zhao}]{peng2021hyperbolic}
Wei Peng, Tuomas Varanka, Abdelrahman Mostafa, Henglin Shi, and Guoying Zhao.
  2021.
\newblock Hyperbolic deep neural networks: A survey.
\newblock \emph{IEEE Transactions on Pattern Analysis and Machine
  Intelligence}.

\bibitem[{Press et~al.(2022)Press, Smith, and
  Lewis}]{press2022trainshorttestlong}
Ofir Press, Noah~A. Smith, and Mike Lewis. 2022.
\newblock \href {https://arxiv.org/abs/2108.12409} {Train short, test long:
  Attention with linear biases enables input length extrapolation}.
\newblock \emph{Preprint}, arXiv:2108.12409.

\bibitem[{Rae et~al.(2019)Rae, Potapenko, Jayakumar, and
  Lillicrap}]{rae2019compressivetransformerslongrangesequence}
Jack~W. Rae, Anna Potapenko, Siddhant~M. Jayakumar, and Timothy~P. Lillicrap.
  2019.
\newblock \href {https://arxiv.org/abs/1911.05507} {Compressive transformers
  for long-range sequence modelling}.
\newblock \emph{Preprint}, arXiv:1911.05507.

\bibitem[{Raffel et~al.(2023)Raffel, Shazeer, Roberts, Lee, Narang, Matena,
  Zhou, Li, and Liu}]{raffel2023exploringlimitstransferlearning}
Colin Raffel, Noam Shazeer, Adam Roberts, Katherine Lee, Sharan Narang, Michael
  Matena, Yanqi Zhou, Wei Li, and Peter~J. Liu. 2023.
\newblock \href {https://arxiv.org/abs/1910.10683} {Exploring the limits of
  transfer learning with a unified text-to-text transformer}.
\newblock \emph{Preprint}, arXiv:1910.10683.

\bibitem[{Ruoss et~al.(2023)Ruoss, Delétang, Genewein, Grau-Moya, Csordás,
  Bennani, Legg, and Veness}]{ruoss2023randomizedpositionalencodingsboost}
Anian Ruoss, Grégoire Delétang, Tim Genewein, Jordi Grau-Moya, Róbert
  Csordás, Mehdi Bennani, Shane Legg, and Joel Veness. 2023.
\newblock \href {https://arxiv.org/abs/2305.16843} {Randomized positional
  encodings boost length generalization of transformers}.
\newblock \emph{Preprint}, arXiv:2305.16843.

\bibitem[{Sarkar(2011)}]{sarkar2011low}
Rik Sarkar. 2011.
\newblock Low distortion delaunay embedding of trees in hyperbolic plane.
\newblock In \emph{International Symposium on Graph Drawing}, pages 355--366.
  Springer.

\bibitem[{Shaham et~al.(2022)Shaham, Segal, Ivgi, Efrat, Yoran, Haviv, Gupta,
  Xiong, Geva, Berant, and Levy}]{shaham2022scrollsstandardizedcomparisonlong}
Uri Shaham, Elad Segal, Maor Ivgi, Avia Efrat, Ori Yoran, Adi Haviv, Ankit
  Gupta, Wenhan Xiong, Mor Geva, Jonathan Berant, and Omer Levy. 2022.
\newblock \href {https://arxiv.org/abs/2201.03533} {Scrolls: Standardized
  comparison over long language sequences}.
\newblock \emph{Preprint}, arXiv:2201.03533.

\bibitem[{Shaw et~al.(2018)Shaw, Uszkoreit, and
  Vaswani}]{shaw2018selfattentionrelativepositionrepresentations}
Peter Shaw, Jakob Uszkoreit, and Ashish Vaswani. 2018.
\newblock \href {https://arxiv.org/abs/1803.02155} {Self-attention with
  relative position representations}.
\newblock \emph{Preprint}, arXiv:1803.02155.

\bibitem[{Song et~al.(2022{\natexlab{a}})Song, Feng, and Jing}]{hyper1}
Mingyang Song, Yi~Feng, and Liping Jing. 2022{\natexlab{a}}.
\newblock \href {https://doi.org/10.18653/V1/2022.NAACL-MAIN.419} {Hyperbolic
  relevance matching for neural keyphrase extraction}.
\newblock In \emph{Proceedings of the 2022 Conference of the North American
  Chapter of the Association for Computational Linguistics: Human Language
  Technologies, {NAACL} 2022, Seattle, WA, United States, July 10-15, 2022},
  pages 5710--5720. Association for Computational Linguistics.

\bibitem[{Song et~al.(2022{\natexlab{b}})Song, Feng, and Jing}]{hyper4}
Mingyang Song, Yi~Feng, and Liping Jing. 2022{\natexlab{b}}.
\newblock \href {https://doi.org/10.1145/3511808.3557538} {A preliminary
  exploration of extractive multi-document summarization in hyperbolic space}.
\newblock In \emph{Proceedings of the 31st {ACM} International Conference on
  Information {\&} Knowledge Management, Atlanta, GA, USA, October 17-21,
  2022}, pages 4505--4509. {ACM}.

\bibitem[{Song et~al.(2023{\natexlab{a}})Song, Feng, and Jing}]{hyper3}
Mingyang Song, Yi~Feng, and Liping Jing. 2023{\natexlab{a}}.
\newblock \href {https://doi.org/10.1145/3543507.3583197} {Hisum: Hyperbolic
  interaction model for extractive multi-document summarization}.
\newblock In \emph{Proceedings of the {ACM} Web Conference 2023, {WWW} 2023,
  Austin, TX, USA, 30 April 2023 - 4 May 2023}, pages 1427--1436. {ACM}.

\bibitem[{Song et~al.(2023{\natexlab{b}})Song, Liu, Feng, and Jing}]{hyper2}
Mingyang Song, Huafeng Liu, Yi~Feng, and Liping Jing. 2023{\natexlab{b}}.
\newblock \href {https://doi.org/10.18653/V1/2023.FINDINGS-ACL.66} {Improving
  embedding-based unsupervised keyphrase extraction by incorporating structural
  information}.
\newblock In \emph{Findings of the Association for Computational Linguistics:
  {ACL} 2023, Toronto, Canada, July 9-14, 2023}, pages 1041--1048. Association
  for Computational Linguistics.

\bibitem[{Song et~al.(2023{\natexlab{c}})Song, Liu, and Jing}]{hyper5}
Mingyang Song, Huafeng Liu, and Liping Jing. 2023{\natexlab{c}}.
\newblock \href {https://doi.org/10.18653/V1/2023.EMNLP-MAIN.997} {Hyperrank:
  Hyperbolic ranking model for unsupervised keyphrase extraction}.
\newblock In \emph{Proceedings of the 2023 Conference on Empirical Methods in
  Natural Language Processing, {EMNLP} 2023, Singapore, December 6-10, 2023},
  pages 16070--16080. Association for Computational Linguistics.

\bibitem[{Su et~al.(2023)Su, Lu, Pan, Murtadha, Wen, and
  Liu}]{su2023roformerenhancedtransformerrotary}
Jianlin Su, Yu~Lu, Shengfeng Pan, Ahmed Murtadha, Bo~Wen, and Yunfeng Liu.
  2023.
\newblock \href {https://arxiv.org/abs/2104.09864} {Roformer: Enhanced
  transformer with rotary position embedding}.
\newblock \emph{Preprint}, arXiv:2104.09864.

\bibitem[{Sun et~al.(2022)Sun, Dong, Patra, Ma, Huang, Benhaim, Chaudhary,
  Song, and Wei}]{sun2022lengthextrapolatabletransformer}
Yutao Sun, Li~Dong, Barun Patra, Shuming Ma, Shaohan Huang, Alon Benhaim,
  Vishrav Chaudhary, Xia Song, and Furu Wei. 2022.
\newblock \href {https://arxiv.org/abs/2212.10554} {A length-extrapolatable
  transformer}.
\newblock \emph{Preprint}, arXiv:2212.10554.

\bibitem[{Suzuki et~al.(2021{\natexlab{a}})Suzuki, Nitanda, Wang, Xu,
  Yamanishi, and Cavazza}]{suzuki2021generalization1}
Atsushi Suzuki, Atsushi Nitanda, Jing Wang, Linchuan Xu, Kenji Yamanishi, and
  Marc Cavazza. 2021{\natexlab{a}}.
\newblock Generalization error bound for hyperbolic ordinal embedding.
\newblock In \emph{International Conference on Machine Learning}, pages
  10011--10021. PMLR.

\bibitem[{Suzuki et~al.(2021{\natexlab{b}})Suzuki, Nitanda, Xu, Yamanishi,
  Cavazza et~al.}]{suzuki2021generalization2}
Atsushi Suzuki, Atsushi Nitanda, Linchuan Xu, Kenji Yamanishi, Marc Cavazza,
  et~al. 2021{\natexlab{b}}.
\newblock Generalization bounds for graph embedding using negative sampling:
  Linear vs hyperbolic.
\newblock \emph{Advances in Neural Information Processing Systems},
  34:1243--1255.

\bibitem[{Touvron et~al.(2023)Touvron, Martin, Stone, Albert, Almahairi,
  Babaei, Bashlykov, Batra, Bhargava, Bhosale, Bikel, Blecher, Ferrer, Chen,
  Cucurull, Esiobu, Fernandes, Fu, Fu, Fuller, Gao, Goswami, Goyal, Hartshorn,
  Hosseini, Hou, Inan, Kardas, Kerkez, Khabsa, Kloumann, Korenev, Koura,
  Lachaux, Lavril, Lee, Liskovich, Lu, Mao, Martinet, Mihaylov, Mishra,
  Molybog, Nie, Poulton, Reizenstein, Rungta, Saladi, Schelten, Silva, Smith,
  Subramanian, Tan, Tang, Taylor, Williams, Kuan, Xu, Yan, Zarov, Zhang, Fan,
  Kambadur, Narang, Rodriguez, Stojnic, Edunov, and
  Scialom}]{touvron2023llama2openfoundation}
Hugo Touvron, Louis Martin, Kevin Stone, Peter Albert, Amjad Almahairi, Yasmine
  Babaei, Nikolay Bashlykov, Soumya Batra, Prajjwal Bhargava, Shruti Bhosale,
  Dan Bikel, Lukas Blecher, Cristian~Canton Ferrer, Moya Chen, Guillem
  Cucurull, David Esiobu, Jude Fernandes, Jeremy Fu, Wenyin Fu, Brian Fuller,
  Cynthia Gao, Vedanuj Goswami, Naman Goyal, Anthony Hartshorn, Saghar
  Hosseini, Rui Hou, Hakan Inan, Marcin Kardas, Viktor Kerkez, Madian Khabsa,
  Isabel Kloumann, Artem Korenev, Punit~Singh Koura, Marie-Anne Lachaux,
  Thibaut Lavril, Jenya Lee, Diana Liskovich, Yinghai Lu, Yuning Mao, Xavier
  Martinet, Todor Mihaylov, Pushkar Mishra, Igor Molybog, Yixin Nie, Andrew
  Poulton, Jeremy Reizenstein, Rashi Rungta, Kalyan Saladi, Alan Schelten, Ruan
  Silva, Eric~Michael Smith, Ranjan Subramanian, Xiaoqing~Ellen Tan, Binh Tang,
  Ross Taylor, Adina Williams, Jian~Xiang Kuan, Puxin Xu, Zheng Yan, Iliyan
  Zarov, Yuchen Zhang, Angela Fan, Melanie Kambadur, Sharan Narang, Aurelien
  Rodriguez, Robert Stojnic, Sergey Edunov, and Thomas Scialom. 2023.
\newblock \href {https://arxiv.org/abs/2307.09288} {Llama 2: Open foundation
  and fine-tuned chat models}.
\newblock \emph{Preprint}, arXiv:2307.09288.

\bibitem[{Vaswani et~al.(2023)Vaswani, Shazeer, Parmar, Uszkoreit, Jones,
  Gomez, Kaiser, and Polosukhin}]{vaswani2023attentionneed}
Ashish Vaswani, Noam Shazeer, Niki Parmar, Jakob Uszkoreit, Llion Jones,
  Aidan~N. Gomez, Lukasz Kaiser, and Illia Polosukhin. 2023.
\newblock \href {https://arxiv.org/abs/1706.03762} {Attention is all you need}.
\newblock \emph{Preprint}, arXiv:1706.03762.

\bibitem[{Wang and Chen(2020)}]{wang2020positionembeddingslearnempirical}
Yu-An Wang and Yun-Nung Chen. 2020.
\newblock \href {https://arxiv.org/abs/2010.04903} {What do position embeddings
  learn? an empirical study of pre-trained language model positional encoding}.
\newblock \emph{Preprint}, arXiv:2010.04903.

\bibitem[{Xiong et~al.(2023)Xiong, Liu, Molybog, Zhang, Bhargava, Hou, Martin,
  Rungta, Sankararaman, Oguz, Khabsa, Fang, Mehdad, Narang, Malik, Fan,
  Bhosale, Edunov, Lewis, Wang, and
  Ma}]{xiong2023effectivelongcontextscalingfoundation}
Wenhan Xiong, Jingyu Liu, Igor Molybog, Hejia Zhang, Prajjwal Bhargava, Rui
  Hou, Louis Martin, Rashi Rungta, Karthik~Abinav Sankararaman, Barlas Oguz,
  Madian Khabsa, Han Fang, Yashar Mehdad, Sharan Narang, Kshitiz Malik, Angela
  Fan, Shruti Bhosale, Sergey Edunov, Mike Lewis, Sinong Wang, and Hao Ma.
  2023.
\newblock \href {https://arxiv.org/abs/2309.16039} {Effective long-context
  scaling of foundation models}.
\newblock \emph{Preprint}, arXiv:2309.16039.

\bibitem[{Yang et~al.(2021)Yang, Chen, Li, Yu, and Xu}]{yang2021hyper}
Haoran Yang, Hongxu Chen, Lin Li, Philip~S Yu, and Guandong Xu. 2021.
\newblock Hyper meta-path contrastive learning for multi-behavior
  recommendation.
\newblock \emph{arXiv preprint arXiv:2109.02859}.

\bibitem[{Yang et~al.(2022)Yang, Li, Zhou, Liu, and King}]{yang2022hicf}
Menglin Yang, Zhihao Li, Min Zhou, Jiahong Liu, and Irwin King. 2022.
\newblock {HICF}: Hyperbolic informative collaborative filtering.
\newblock In \emph{Proceedings of the 28th ACM SIGKDD Conference on Knowledge
  Discovery and Data Mining}, pages 2212--2221.

\bibitem[{Zhong et~al.(2021)Zhong, Yin, Yu, Zaidi, Mutuma, Jha, Awadallah,
  Celikyilmaz, Liu, Qiu, and Radev}]{zhong2021qmsumnewbenchmarkquerybased}
Ming Zhong, Da~Yin, Tao Yu, Ahmad Zaidi, Mutethia Mutuma, Rahul Jha,
  Ahmed~Hassan Awadallah, Asli Celikyilmaz, Yang Liu, Xipeng Qiu, and Dragomir
  Radev. 2021.
\newblock \href {https://arxiv.org/abs/2104.05938} {Qmsum: A new benchmark for
  query-based multi-domain meeting summarization}.
\newblock \emph{Preprint}, arXiv:2104.05938.

\bibitem[{Zhou et~al.(2022)Zhou, Li, Yang, and Pan}]{zhou2022telegraph}
Min Zhou, Bisheng Li, Menglin Yang, and Lujia Pan. 2022.
\newblock Telegraph: A benchmark dataset for hierarchical link prediction.
\newblock \emph{arXiv preprint arXiv:2204.07703}.

\end{thebibliography}
\section{Appendix}
\subsection{Proof}
\textbf{Theorem} \ref{thm:1} (Positional Discrimination Capacity)
For any relative position $r \in \mathbb{Z}$ and query vector $q, ~ \exists$ key vector $k$ such that:
\begin{equation}
\underset{s}{\mathrm{argmax}}(\langle f_q(m), f_k(m+s) \rangle) = r
\end{equation}

Frist, considering the 2-dimensional case, the attention weight can be calculated as:
\begin{align}
V(q_m,k_n) &= \boldsymbol{q}_m^T
B(\theta,m)
B'(\theta,n)
\boldsymbol{k}_n
\end{align}
if
\begin{equation}
\underset{s}{\mathrm{argmax}}(\langle f_q(m), f_k(m+s) \rangle) = t \neq r
\end{equation}
which means
\begin{align}
V(q_m,k_t) &>V(q_m,k_r)
\end{align}
such that
\begin{align}
\boldsymbol{q}_m^T
B(\theta,m)
B'(\theta,t)
\boldsymbol{k}_t>
\boldsymbol{q}_m^T
B(\theta,m)
B'(\theta,r)
\boldsymbol{k}_r
\end{align}
We only need to reconstruct $\boldsymbol{k}_r$ so that 
\begin{align}
\boldsymbol{k}_r>\frac{B'(\theta,t)}{B'(\theta,r)}\boldsymbol{k}_t
\end{align}
So,in 2-dimension case,$\exists$ key vector $k$ such that:
\begin{equation}
\underset{s}{\mathrm{argmax}}(\langle f_q(m), f_k(m+s) \rangle) = r
\end{equation}
When generalized to 2-n dimensions, the theorem maintains its original properties. By constructing k in this way for every 2-dimensional subspace, it ensures that
\begin{align}
\underset{s}{\mathrm{argmax}}
   \sum_{k=1}^{n} \left( \mathbf{q}_m^{(k)} \right)^{\top} \rho(g_k)^{m-n} \mathbf{k}_n^{(k)} =r
\end{align}

\subsection{ Analysis of RoPE and HoPE}
\subsubsection{RoPE}
For simplicity of notation in this work, we follow the assumption \cite{barbero2024roundroundgomakes} that queries and keys are $d$ dimensional vectors with $d \geq 2$ being an even number. We decompose queries and keys into 2-dimensional chunks $\mathbf{q}_i = \bigoplus_{k=1,\ldots,d/2} \mathbf{q}_i^{[k,k+1]} = \bigoplus_{k=1,\ldots,d/2} \mathbf{q}_i^{(k)}$, where $\bigoplus$ denotes direct sum (concatenation). In other words, we denote by $\mathbf{q}_i^{(k)} \in \mathbb{R}^2$ the $k$-th 2-dimensional chunk of the query vector of the $i$-th token, using analogous notation for the key vectors.

RoPE considers a sequence of angles $G = \left(g_k = \theta^{-2(k-1)/d}: k = 1, \ldots, d/2\right)^2$, where $g_1 = 1$ is the fastest rotating component at 1 radian per token and $g_{d/2} = \theta^{-(d-2)/d} \approx \theta^{-1}$ the slowest rotating component at approximately $1/\theta$ rotations per token. The parameter $\theta$ is called the base wavelength, which by default is 10,000. We denote by $\rho(g_k)$ the matrix form of $g_k$:
\begin{equation}
\rho(g_k) = \begin{bmatrix}
\cos(g_k) & -\sin(g_k) \\
\sin(g_k) & \cos(g_k)
\end{bmatrix},
\end{equation}
highlighting that $\rho(g_k)$ is a 2-dimensional orthogonal transformation (rotation). One can view $\rho(g_k)$ as a `unit rotation' by $g_k$ radians. The RoPE technique amounts to the construction of a block-diagonal matrix $\mathbf{R}^i = \bigoplus_{k=1,\ldots,d/2} \rho(g_k)^i \in \mathbb{R}^{d \times d}$, where each $2 \times 2$ block on the diagonal is a rotation by a different frequency of RoPE. The $\mathbf{R}^i$ denotes in fact matrix exponentiation by an integer $i$, which is the position of $\mathbf{x}_i$. We can exploit a nice property of rotation matrices, i.e., that $\rho(g_k)^i = \rho(i g_k)$ to avoid the computation of the matrix power. As this matrix is block diagonal, computing $\mathbf{R}_i \mathbf{q}_i$ means that the rotations act only on 2-dimensional chunks of the query (or key), i.e., $\mathbf{R}_i \mathbf{q}_i = \bigoplus_{k=1,\ldots,d/2} \rho(i g_k) \mathbf{q}_i^{(k)}$. This leads to the final formulation of $k_{\text{RoPE}}$:

\begin{equation}
\begin{aligned}
k_{\text{RoPE}}(\mathbf{q}_i, \mathbf{k}_j) &= (\mathbf{R}^i \mathbf{q}_i)^\top (\mathbf{R}^j \mathbf{k}_j) \\
&= \mathbf{q}_i^\top \mathbf{R}^{j-i} \mathbf{k}_j \\
&= \sum_{k=1,\ldots,d/2} \left(\mathbf{q}_i^{(k)}\right)^\top \rho(g_k)^{j-i} \mathbf{k}_j^{(k)}
\end{aligned}
\end{equation}
where we use the fact that $\left(\rho(g_k)^i\right)^\top \rho(g_k)^j = \rho(g_k)^{-i} \rho(g_k)^j = \rho(g_k)^{j-i}$. We highlight how the block diagonal structure of $\mathbf{R}$ allows one to decompose the dot product into the sum of dot products of 2-dimensional chunks, with each key vector chunk rotated at a frequency dictated by $g_k$.
\\
Considering the smallest and largest values of $\theta$ in RoPE , denoted as $\theta_{\text{min}}$ and $\theta_{\text{max}}$, respectively, and their corresponding wavelengths, $\lambda_{\text{max}}$ and $\lambda_{\text{min}}$. When a sequence is input into the model, it essentially performs an uneven positional encoding on the sequence. 

When the sequence length begins to exceed $\lambda_{\text{min}}$, dimensions start completing one cycle of rotation. As the sequence length continues to increase, more dimensions reach their rotational cycles. This phenomenon endows RoPE with the ability to extrapolate. When the sequence length increases, some dimensions' rotational values have been seen during the model's previous training, providing familiarity for longer sequences. However, this also introduces noise into the training process. As the sequence length increases, the attention weights ($\text{attn\_weight}$) on these dimensions may become larger than those for shorter relative positions, confusing the model. This issue is difficult to mitigate by merely adjusting the frequency.
\begin{figure}[ht]
    \centering 
    \includegraphics[width=0.49\textwidth]{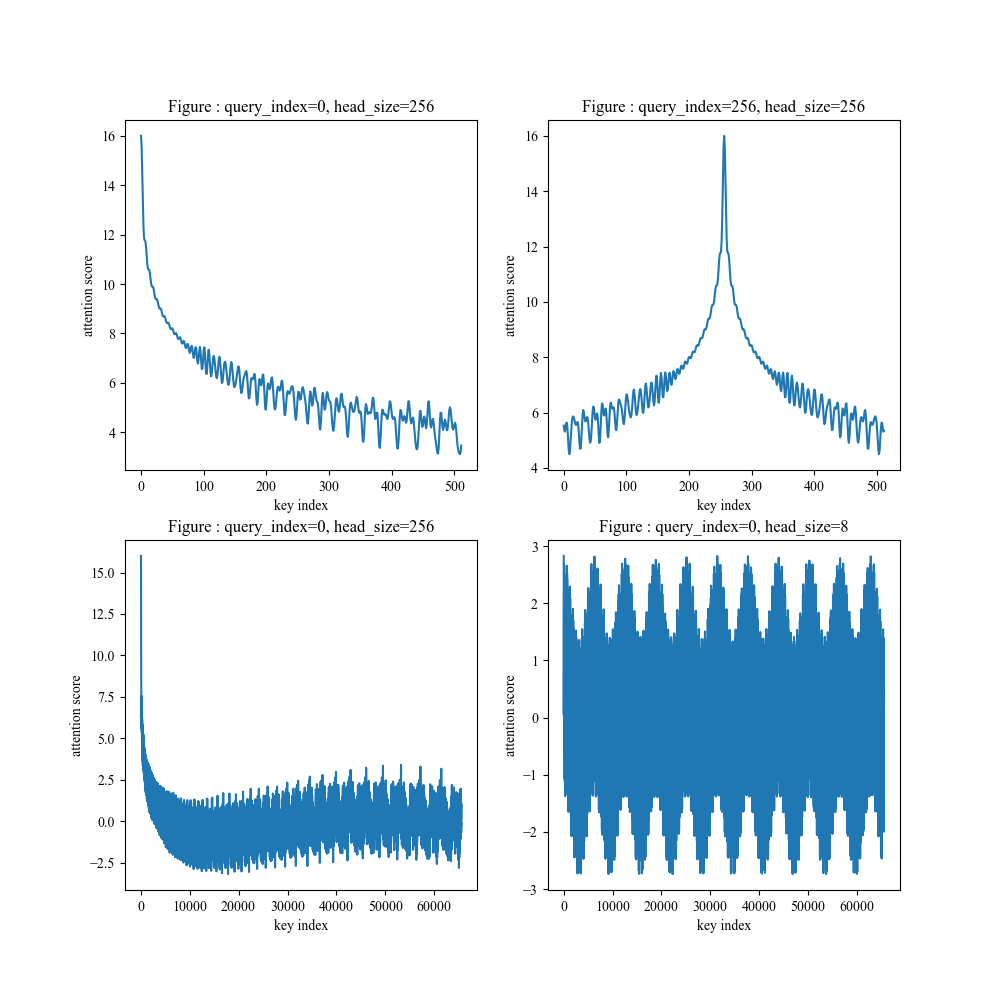} 
    \caption{Attention weight decay trend with rope } 
    \label{fig:rope_trend_4} 
\end{figure}
The extrapolation capability of RoPE arises from this characteristic. In extreme cases, if the sequence length exceeds the lowest frequency, i.e., the longest wavelength $\lambda_{\text{max}}$, RoPE degenerates into no position encoding, failing to provide positional information to the model anymore. 

\subsubsection{HoPE}
By substituting the rotation matrix of the RoPE (Rotary Position Embedding) with a hyperbolic rotation matrix, we characterize the relative information by rotating $q$ by $m\theta$ and $k$ by $-m\theta$. This approach employs hyperbolic matrices to encode the positional information differently, capturing the relative positions between tokens effectively. Consistent with the matrix form mentioned above, we can derive the following matrix.
\begin{equation}
\rho(g_k) = \begin{bmatrix}
\cosh(g_k) & \sinh(g_k) \\
\sinh(g_k) & \cosh(g_k)
\end{bmatrix},
\end{equation}
However, the rotation matrix of RoPE is an orthogonal matrix, which means it will not change the modulus of any vector:
\begin{equation}
\begin{aligned}
&\begin{bmatrix}
\cos(\theta) & -\sin(\theta) \\
\sin(\theta) & \cos(\theta)
\end{bmatrix}
\begin{bmatrix}
q_1 \\
q_2
\end{bmatrix}
\\
&=
\begin{bmatrix}
q_1\cos(\theta) - q_2\sin(\theta) \\
q_1\sin(\theta) + q_2\cos(\theta)
\end{bmatrix}
\end{aligned}
\end{equation}
\text{The length after rotation remains unchanged.}
However, our hyperbolic rotation is not an orthogonal matrix, and it will change the modulus of the vector.
\begin{equation}
\begin{aligned}
&\begin{bmatrix}
\cosh(\theta) & \sinh(\theta) \\
\sinh(\theta) & \cosh(\theta)
\end{bmatrix}
\begin{bmatrix}
q_1 \\
q_2
\end{bmatrix}
\\
&=
\begin{bmatrix}
q_1\cosh(\theta) + q_2\sinh(\theta) \\
q_1\sinh(\theta) + q_2\cosh(\theta)
\end{bmatrix}
\end{aligned}
\end{equation}
\text{Original length: }
\begin{equation}
 \sqrt{q_1^2 + q_2^2} = A
\end{equation}
\text{Rotated length: }
\begin{equation}
\begin{aligned}
\sqrt{(q_1^2 + q_2^2)\cosh(2\theta) + 2q_1q_2\sinh(2\theta)} = B
\end{aligned}
\end{equation}
To find the relationship between A and B, we got B divided by A
\begin{equation}
\frac{B}{A} = \sqrt{\cosh(2\theta) + \frac{2q_1q_2\sinh(2\theta)}{q_1^2 + q_2^2}}
\end{equation}
According to the inequality related to the binomial theorem: $a^2 + b^2 \geq 2ab$.
\begin{equation}
\begin{aligned}
\frac{B}{A} &\leq \sqrt{e^{2\theta}}
\\
&= e^{\theta}
\end{aligned}
\end{equation}
Therefore, to account for the corresponding multiplication of the modulus change, we premultiply the hyperbolic rotation matrix by a penalty coefficient.
\begin{equation}
e^{-\theta}
\begin{bmatrix}
\cosh(\theta) & \sinh(\theta) \\
\sinh(\theta) & \cosh(\theta)
\end{bmatrix} 
\end{equation}
This approach not only reduces the noise during training but also preserves the assumptions of positional encoding.
\subsection{ Experimental Details}
\subsubsection{Perplexity Experiment}
\textbf{Model configurations.} In this experiment, we train decoder-only Transformer language models with different positional encoding techniques while keeping all the other configurations the same. For RoPE, we follow \cite{su2023roformerenhancedtransformerrotary} to set the hyperparameters in the rotary matrix, respectively. For ALiBi, we follow \cite{press2022trainshorttestlong} to set the slope values in each attention head. For the intra-segment encoding of  BiPE, we use the learnable absolute positional encoding. For the inter-segment encoding of  BiPE-RoPE, the hyperparameters are kept the same as the original setting. For the inter-segment encoding of our BiPE-ALiBi, the slope values are set to 96 times of the original ALiBi's setting. Other model configurations are provided in Table \ref{tab:4}.

\begin{table}[h]
\centering
\caption{Model configurations for length extrapolation.}
\begin{tabular}{cc}
\hline \hline
Layers & 12 \\
Attention heads & 12 \\
Head dimensions & 64 \\
Hidden dimensions & 768 \\
FFN dimensions & 3072 \\
Model parameters & 155M \\
\hline \hline
\end{tabular}
\label{tab:4}
\end{table}

\noindent \textbf{Training recipes.} The next token prediction objective is adopted for language model training. All models are trained on the Pile dataset with a total sequence length of 1024. The training recipes are shown in Table \ref{tab:5}.

\begin{table}[h]
\centering
\caption{Training recipes for length extrapolation.}
\begin{tabular}{cc}
\hline \hline
Batch size & 256 \\
Total training epochs & 1 \\
Dropout & 0.0 \\
Weight decay & 0.01 \\
Optimizer & AdamW \\
Learning rate & $1e-4$ \\
\hline \hline
\end{tabular}
\label{tab:5}
\end{table}

\subsubsection{Fine-Tuning Experiment}
\textbf{Fine-tuning on SCROLLS.} We fine-tune pretrained language models with different positional encoding methods on SCROLLS\cite{shaham2022scrollsstandardizedcomparisonlong}. It is a long context benchmark that consists of seven distinct datasets covering different tasks, e.g, Question-Answering (Qasper\cite{dasigi2021datasetinformationseekingquestionsanswers}, NarrativeQA\cite{kočiský2017narrativeqareadingcomprehensionchallenge}, and QuALITY\cite{pang2022qualityquestionansweringlong}), Natural Language Inference (ContractNLI\cite{koreeda2021contractnlidatasetdocumentlevelnatural}) and Summarization (QMSum\cite{zhong2021qmsumnewbenchmarkquerybased}, SummScreenFD\cite{chen2022summscreendatasetabstractivescreenplay}, and GovReport\cite{huang2021efficientattentionslongdocument}). All the model configurations are the same as those in Table \ref{tab:4}.
\begin{table}[h]
\centering
\caption{Finetuning recipes for long context benchmark.}
\begin{tabular}{l c}
\hline \hline
Batch size & 64 \\
Total training steps & 5000 \\
Dropout & 0.0 \\
Weight decay & 0.01 \\
Optimizer & AdamW \\
Learning rate & $1e-5$ \\
\hline \hline
\end{tabular}
\label{tab:6}
\end{table}
\\
\textbf{Fine-tuning recipes.} We fine-tune models using the next token prediction objective on each task with a sequence length of 8192. The fine-tuning recipes are provided in Table \ref{tab:6}
\subsection{Lorentz Transformation And Lorentz Group}

The general form of the Lorentz transformation can be written as:
$$
x'^\mu = \Lambda^\mu_{\ \nu} x^\nu
$$
where $\Lambda^\mu_{\ \nu}$ is the Lorentz transformation matrix satisfying the condition:
$$
\eta_{\rho\sigma} \Lambda^\rho_{\ \mu} \Lambda^\sigma_{\ \nu} = \eta_{\mu\nu}
$$
with $\eta_{\mu\nu}$ being the Minkowski metric:
$$
\eta_{\mu\nu} = \text{diag}(1, -1, -1, -1)
$$

For a specific case of a boost along the $x$-axis, the transformations are given by:
\begin{align*}
t' &= \gamma \left(t - \frac{vx}{c^2}\right) \\
x' &= \gamma (x - vt) \\
y' &= y \\
z' &= z
\end{align*}
where $\gamma = \frac{1}{\sqrt{1-\frac{v^2}{c^2}}}$ is the Lorentz factor.

The set of all such transformations forms the Lorentz group, denoted as $\mathrm{O}(1,3)$, which is defined as:
$$
\underbrace{\mathrm{O}(1,3) \equiv \left\{ \Lambda \mid \Lambda \in \mathrm{GL}(4,\mathbb{R}), g_{\mu\nu} \Lambda^\mu{}_\rho \Lambda^\nu{}_\sigma = g_{\rho\sigma} \right\}}_{\dim \mathrm{O}(1,3)=6}
$$

This group includes rotations and boosts (velocity transformations) and has six degrees of freedom: three for rotations and three for boosts. It represents the fundamental symmetry group of special relativity, ensuring the invariance of physical laws across different inertial frames.

Essentially, $\mathrm{O}(1,3)$ is the group of linear transformations on Minkowski spacetime that preserve the metric. For more details on metric spaces and index notation, refer to Masaki Notation.

\begin{algorithm}[t]
\SetAlgoLined
\KwIn{q, k: [batch, head, seq, dim]; theta: [dim//2] (per-dimension angle); theta\_prime}
\KwOut{Modified q and k}
\For{$i \gets 0$ \textbf{to} $\text{dim}-2$ \textbf{step} 2}{
    $\text{angle} \gets \text{pos} \cdot \theta[i/2]$\;
    $c, s \gets \cosh(\text{angle}), \sinh(\text{angle})$\;
    $\text{rot\_q} \gets [c \cdot q[...,i] + s \cdot q[...,i+1],\ s \cdot q[...,i] + c \cdot q[...,i+1]]$\;
    $\text{rot\_k} \gets [c \cdot k[...,i] - s \cdot k[...,i+1],\ -s \cdot k[...,i] + c \cdot k[...,i+1]]$\;
    $q[...,i:i+2] \gets \exp(-\text{pos} \cdot \theta_{\text{prime}}) \cdot \text{rot\_q}$\;
    $k[...,i:i+2] \gets \exp(\text{pos} \cdot \theta_{\text{prime}}) \cdot \text{rot\_k}$\;
}
\caption{HoPE}
\end{algorithm}
\end{document}